\documentclass{article}
\usepackage{hyperref}
\usepackage{spconf,amsmath,graphicx}
\usepackage[ruled,vlined]{algorithm2e}
\usepackage{amsmath}
\usepackage{halloweenmath}
\usepackage{amsfonts}
\DeclareMathOperator*{\argmax}{arg\,max}
\DeclareMathOperator*{\sign}{sign}
\usepackage{xcolor}
\usepackage[usestackEOL]{stackengine}

\title{WITCHcraft: Efficient PGD attacks with random step size}
\name{\Longstack{Ping-Yeh Chiang$^{\mathwitch}$ \, Jonas Geiping$^{\mathghost}$ \, Micah Goldblum$^{\mathwitch}$ \\
Tom Goldstein$^{\mathwitch}$ \, Renkun Ni$^{\mathwitch}$ \, Steven Reich$^{\mathwitch}$ \, Ali Shafahi$^{\mathwitch}$
}
 \thanks{Authors contributed equally and are listed in alphabetical order.}
 }

\address{$^{\mathwitch}$University of Maryland, College Park \\ $^{\mathghost}$University of Siegen}
\begin{document}

\maketitle

\begin{abstract}
State-of-the-art adversarial attacks on neural networks use expensive iterative methods and numerous random restarts from different initial points.  Iterative FGSM-based methods without restarts trade off performance for computational efficiency because they do not adequately explore the image space and are highly sensitive to the choice of step size.  We propose a variant of Projected Gradient Descent (PGD) that uses a random step size to improve performance without resorting to expensive random restarts.  Our method, Wide Iterative Stochastic crafting (WITCHcraft), achieves results superior to the classical PGD attack on the CIFAR-10 and MNIST data sets but without additional computational cost.  This simple modification of PGD makes crafting attacks more economical, which is important in situations like adversarial training where attacks need to be crafted in real time.
\end{abstract}
\section{Introduction}
\par Neural networks trained using stochastic gradient descent (SGD) are easily fooled by \emph{adversarial examples}, small perturbations to inputs that change the output of the network \cite{szegedy2013intriguing}.  Adversarial attacks can expose serious security vulnerabilities in real-world applications such as object detection in self-driving cars \cite{sitawarin2018darts} and classification in medical imaging \cite{finlayson2018adversarial}.  In response to this threat, subsequent work has developed training methods for producing neural networks robust to these attacks \cite{gu2014towards,madry2018towards}.  The back-and-forth between new defenses and adversarial attacks that break them has spawned an array of powerful new attack methods.

\par Among these, typical \emph{untargeted} adversarial attacks operate by maximizing the loss of a neural network with respect to image space, within a small ball surrounding the input, using various optimization algorithms.  \emph{Targeted} attacks, on the other hand, minimize loss on a particular incorrect label.  In the white-box attack setting, an attacker has access to the parameters of the network, while black-box attacks operate by querying the network or transferring attacks computed on other networks.  We focus on the white-box setting, a space which is dominated by optimization methods.

\begin{figure}
\vspace{.5cm}
{\includegraphics[width=8.7cm]{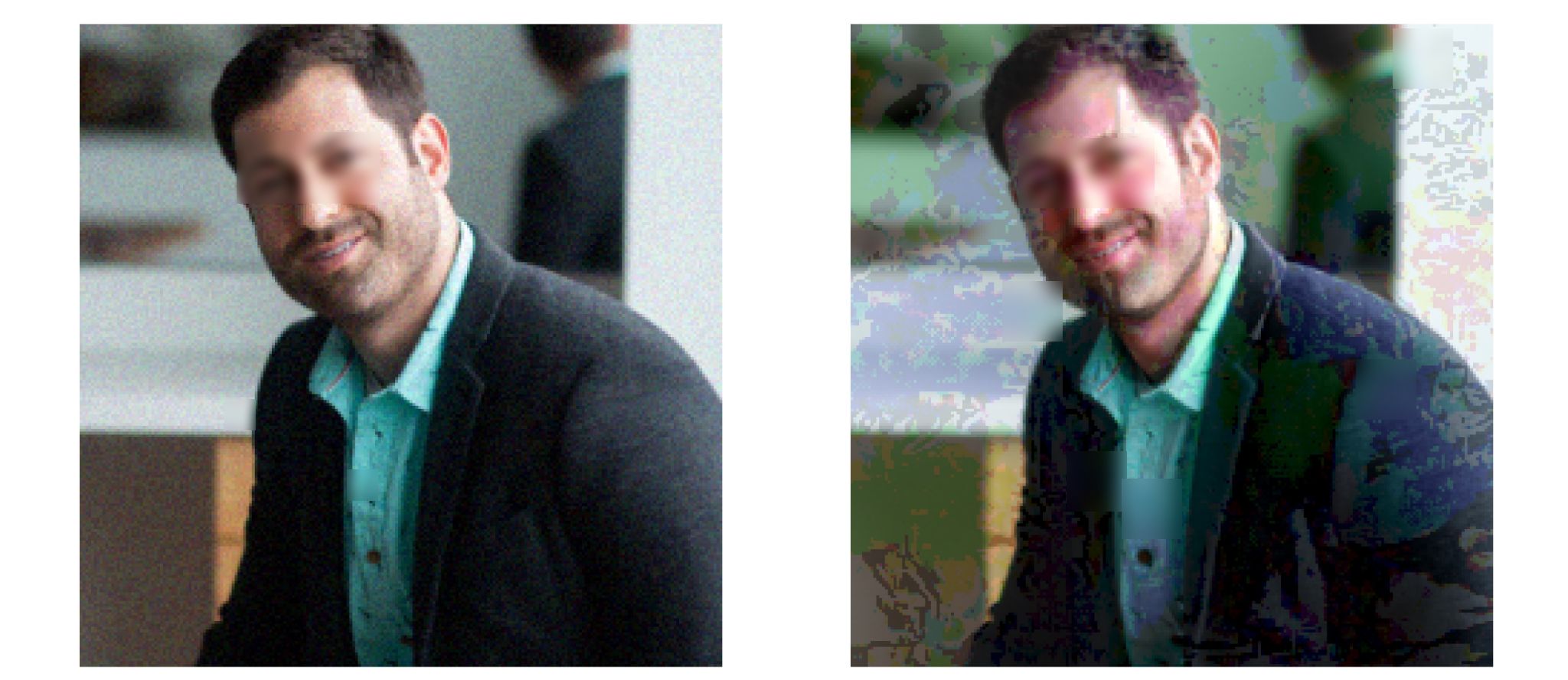}}
\vspace{0cm}
\caption{WITCHcraft perturbs a ``person'' image into the ImageNet ``tench'' class (a kind of fish) using a targeted attack without changing the apparent class to human observers.  This is an example of a ``targeted'' attack because the ``tench'' class was chosen a-priori by the attacker.  The images have been blurred for anonymity.}
\label{figure:tom}
\end{figure}

\par Input spaces in computer vision are high-dimensional, and finding these small perturbations that effectively fool a network requires non-convex optimization \cite{sinha2017certifying}. The outputs of neural networks oscillate in these neighborhoods, so that classical gradient descent is ineffective \cite{athalye2018obfuscated}, and signed gradient descent methods \cite{bernstein2018signsgd} have better success. Even so, a single gradient descent is not guaranteed to solve the problem, so state-of-the-art attacks restart the attack many times with random initialization to introduce randomness and aggressively explore input space. However, this technique increases computational cost which may render an adversary, dynamically attacking a system in real time, useless.  This is particularly problematic for {\em adversarial training}, a process in which attacks are generated on-the-fly during network training and used to harden a network against attacks.

\par In this work, we develop a novel method, Wide Iterative Stochastic crafting (WITCHcraft), for introducing randomness into adversarial attacks without running the attack multiple times with different initializations.  We modify the classical PGD attack, which is similar to the Basic Iterative Method with a random restart and projections, by using a coordinate-wise random step size, meaning each entry of the signed gradient is scaled by a random factor chosen uniformly at random.  We find that this randomization scheme decreases sensitivity to the choice of the step size parameters and initialization.  We compare our method to standard PGD attacks and PGD with random restarts.  We find that our method outperforms both of these attacks on the CIFAR-10 and MNIST data sets when granting all attackers equal compute budget. In Figure \ref{figure:tom}, we see an example of WITCHcraft perturbing a man into an ImageNet fish class without visibly changing the class to human observers \cite{deng2009imagenet}.

\section{Related Work}

Szegedy et al. first demonstrated the existence of adversarial examples comprising small perturbations of input pixels \cite{szegedy2013intriguing}.  Their regularized gradient descent method spawned numerous subsequent attacks.  These later attacks include a variety of new objectives from universal adversarial perturbations, in which a single perturbation is effective on most test images, to realistic perturbations which are not sensitive to certain transformations \cite{moosavi2017universal,athalye2017synthesizing}.  Similarly, defense methods have sprung up to create networks robust to these attacks \cite{gu2014towards,goodfellow2014explaining,ross2018improving}.  The most successful of these defense methods, adversarial training, involves exposing the network to adversarial examples instead of clean examples during training \cite{madry2018towards}.  Since the advent of these defenses, even more attack methods have emerged for defeating adversarially robust networks \cite{brendel2017decision,obfuscated-gradients}.  A recent result of this war between attacks and defenses is the high computational cost of effective attacks against adversarially trained models.

The foundation of most popular attack methods is the PGD attack described by Madry et. al. \cite{madry2018towards}. Their version of this attack starts with a randomly initialized perturbation $\delta \in \mathcal{S}$, which is updated at each step via 
$$ \delta \leftarrow \Pi_{\mathcal{S}} [\delta + \tau \sign( \nabla_\mathbf{x} \mathcal{L}(\mathbf{x} + \delta,y))], $$
where $\tau$ is a fixed step size, $\mathbf{x}$ is the input, and $y$ is the corresponding label. This method uses just the sign of the gradient, a strategy first adapted for attacks in \cite{goodfellow2014explaining}. The superiority of signed gradients to raw gradients for producing adversarial examples has puzzled the robustness community since its discovery, but these strong fluctuations in the gradient signal possibly help the attack to escape suboptimal solutions with low gradient. Signed gradient descent methods are tightly interconnected with adaptive gradient methods, such as Adam \cite{kingma2014adam} as discussed in \cite{bernstein2018signsgd}.

In the aforementioned paper, Madry et al. demonstrate that adversarial training against PGD results in a model that is robust to norm-bounded attacks.  Surprisingly, their experiments (as well as later experiments by other authors \cite{zheng2019distributionally}) show that even though their models are specifically trained against PGD, models adversarially trained against the PGD attacker are also robust against other attacks. 


The current best reported (white-box) attack on the Madry PGD-trained model is the multi-targeted attack described in \cite{qin2019adversarial}, which uses a targeted PGD attack (in which the attacker chooses the label) on each incorrect class to find the best class in which to perturb the clean input. This method exhibits numerical results superior to previous methods but has the drawback of being highly computationally expensive as it both employs random restarts and scales linearly with the number of classes.  This makes it necessary to run the attack on massive servers when training on large data sets with high-resolution images, such as ImageNet or similar.

\section{Our Algorithm}

In our work, we combine the PGD attack with a randomly chosen coordinate-wise step size (See Algorithm \ref{alg:WITCHcraft}).  Effectively, a random step size is chosen independently for each entry in the gradient so that different pixels are perturbed different amounts with each iteration.  WITCHcraft still incorporates a random initialization, which has been found to improve results in previous work \cite{madry2018towards} and comes at no cost to the attack scheme. We terminate the algorithm as soon as the attack is successful at fooling the image classifier.

This strategy of perturbing the gradient signal randomly can be understood as a specific form of stochastic preconditioning of the actual PGD step, which leads to an increasing exploratory power of the optimization scheme. Due to the stochasticity, the algorithm does not easily stagnate or oscillate between two fixed points.  As a result, the method avoids getting trapped in local minima or cycles that inhibit progress. 

Note that the step size in Algorithm \ref{alg:WITCHcraft} is a 2-dimensional or 3-dimensional array of values (the same dimensions as the image being crafted), as opposed to a single scalar value (as is conventionally used for standard PGD attacks).  The step size array $\tau$ is multiplied into the gradient update using a Hadamard (i.e., coordinate-wise) product, denoted $\odot$.  The entries in the step size array are independent and identically distributed and are chosen from uniform distribution on the interval $[0,2a].$ In our experiments, which appear below, we compare this step size choice to a deterministic version with step size $a$, which has an identical expected value to the randomized version.

\SetArgSty{textnormal}
\begin{algorithm}[h!]
$\text{\bf{Requires:}}$ Network $f$, input $\mathbf{x}$, label $y$, permissible perturbation set $\mathcal{S}$, number of steps $n$, and expected step size parameter $a$.\\
 Initialize perturbation $\delta$ with entries distributed independently according to distribution $\mathcal{U}(S)$.\\
  \For{$\text{step}$ = $1$,..., $n$}{
   Sample $\tau$ with entries distributed independently according to distribution $\mathcal{U}(0, 2a)$.\\
   $\delta \leftarrow \Pi_{\mathcal{S}}[\delta + \tau \odot \sign(\nabla_{\mathbf{x}} \mathcal{L}(\mathbf{x}+\delta,\text{class})$\\
   If $\argmax (f(\mathbf{x}+\delta)) \neq y$, return $\mathbf{x}+\delta$ and $\mathbf{break}$.
 }
 \caption{The WITCHcraft attack algorithm.}
 \label{alg:WITCHcraft}
\end{algorithm}

\section{Experiments}

\subsection{Comparison to PGD benchmarks}

\par We test our method on the CIFAR-10 and MNIST data sets against the WideResNet(34-10) model and CNN model with two convolutional layers respectively, trained by the authors of \cite{madry2018towards} using their 7-step PGD adversarial training algorithm \cite{zagoruyko2016wide}.  These robust models are canonical for testing attacks and are used for competitive robustness leaderboards \cite{madry2019github}. We focus on $\ell_{\infty}$ attacks, since this choice of norm dominates the robustness literature.  Perturbations on CIFAR-10 images are restricted to the $\ell_{\infty}$ ball with radius $0.031$, while for MNIST, attacks are restricted to the ball with radius $0.3$.

\begin{table}[h!]
\centering
\vspace{.5cm}
\begin{tabular}{|l|l|}
\hline
Attack  & CIFAR-10 $\mathcal{A}_{adv}$\\ \hline
20-step PGD & 47.04\%  \\ \hline
20-step WITCHcraft & \bf{45.92\%} \\ \hline
100-step PGD & 45.29\%  \\ \hline
100-step WITCHcraft & \bf{45.20\%}  \\ \hline
20-PGD w/ 10 restarts & 45.21\% \\ \hline
\end{tabular}
\vspace{.3cm}
\caption{Robust accuracy, $\mathcal{A}_{adv}$, of various adversarial attacks against the WideResNet(34-10) model trained on CIFAR-10, and released by the authors of \cite{madry2018towards}. Bolded entries indicate best attack results across fixed computational complexity. Randomized coordinate-wise learning rates (WITCHcraft) improve attack effectiveness with a fixed computational budget.}
\label{table:CIFAR10}
\end{table}

\begin{table}[h!]
\centering
\vspace{.5cm}
\begin{tabular}{|l|l|}
\hline
Attack  & MNIST $\mathcal{A}_{adv}$\\ \hline
100-step PGD & 92.52\%  \\ \hline
100-step WITCHcraft & \bf{91.68\%}  \\ \hline
500-step PGD & 91.91\% \\ \hline
500-step WITCHcraft & \bf{91.00\%} \\ \hline
\end{tabular}
\vspace{.3cm}
\caption{Robust accuracy, $\mathcal{A}_{adv}$, of various adversarial attacks against the two-layer CNN model trained on MNIST and released by the authors of \cite{madry2018towards}. Bolded entries indicate the best attack results across fixed computational complexity.  Like we observed for the CIFAR-10 model, randomized coordinate-wise learning rates improve attack effectiveness with a fixed computational budget.}
\label{table:MNIST}
\end{table}

\begin{figure}[htb]
{\includegraphics[width=8.5cm]{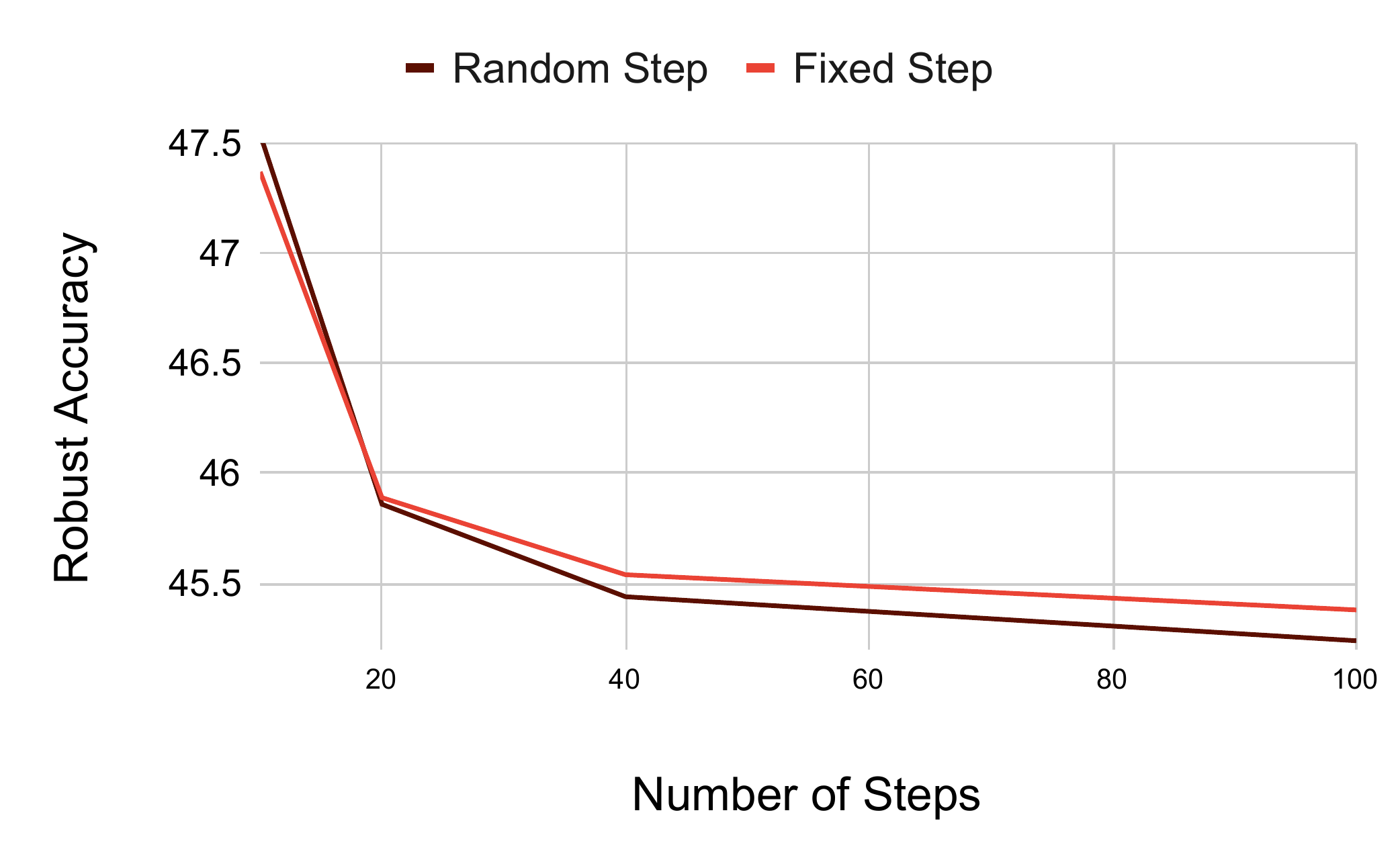}}
\caption{Comparison of robust accuracy as we increase the number of attack steps for WITCHcraft vs. PGD on CIFAR-10. Each reported robust accuracy is an average of 8 trials. As the number of steps increases, WITCHcraft outperforms PGD by a progressively wider margin.}
\vspace{.5cm}
\label{figure:cifarsteps}
\end{figure}

WITCHcraft outperforms PGD with Madry's choice of hyperparameters and the same number of updates as shown in Table \ref{table:CIFAR10} and Table \ref{table:MNIST}. It is especially interesting to note that WITCHcraft is able to continuously improve during the attack iterations, whereas the standard PGD method quickly saturates as shown in Figure \ref{figure:cifarsteps} and Figure \ref{figure:mniststeps}.

\subsection{Exploring the effect of expected step size}

\begin{figure}[htb]
{\includegraphics[width=8.5cm]{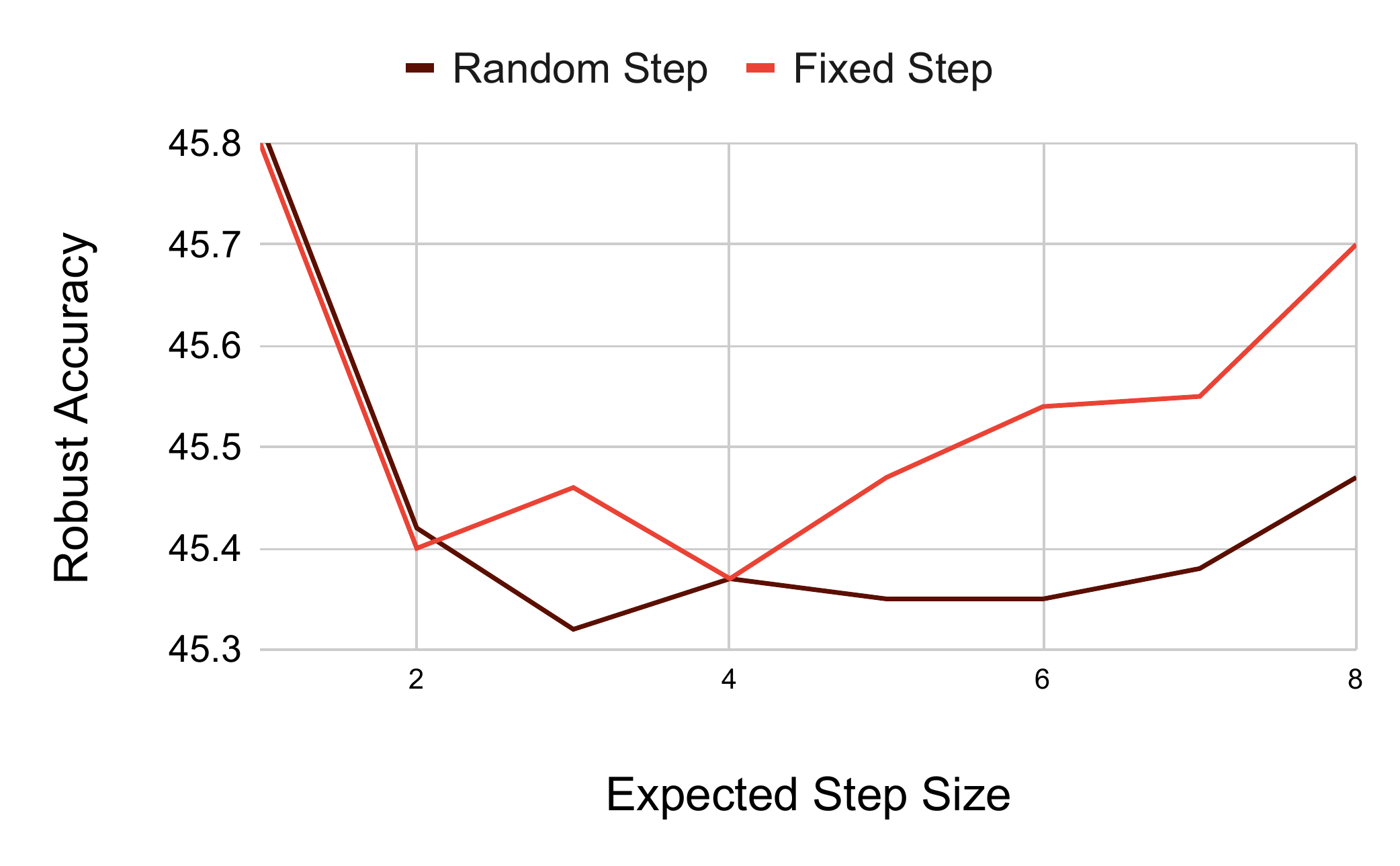}}
\caption{Sensitivity plot of a 40-step PGD attack compared with 40-step WITCHcraft for the CIFAR-10 challenge.  The horizontal axis represents expected step size and the vertical axis represents robust accuracy.  We see that the randomized step size choice in WITCHcraft out-performs a deterministic step size choice, particularly when larger step sizes are used.}
\vspace{.5cm}
\label{figure:sensitivity}
\end{figure}

Following up on this apparent success, we investigate the sensitivity of PGD and WITCHcraft by comparing the expected step size of our approach and the corresponding fixed step size for PGD. The plot in Figure \ref{figure:sensitivity} shows that the standard PGD attack can, in fact, be further enhanced over Madry's results by fine-tuning the step size.  WITCHcraft shows some sensitivity to expected step size but performs at least as well as standard PGD except for very small values of hyperparameters, where randomness seemingly has little to no effect on the optimization. Of particular note is that the best overall reduction in accuracy among these trials is achieved by WITCHcraft at an expected step size of $3$. 

\begin{figure}[htb]
{\includegraphics[width=8.5cm]{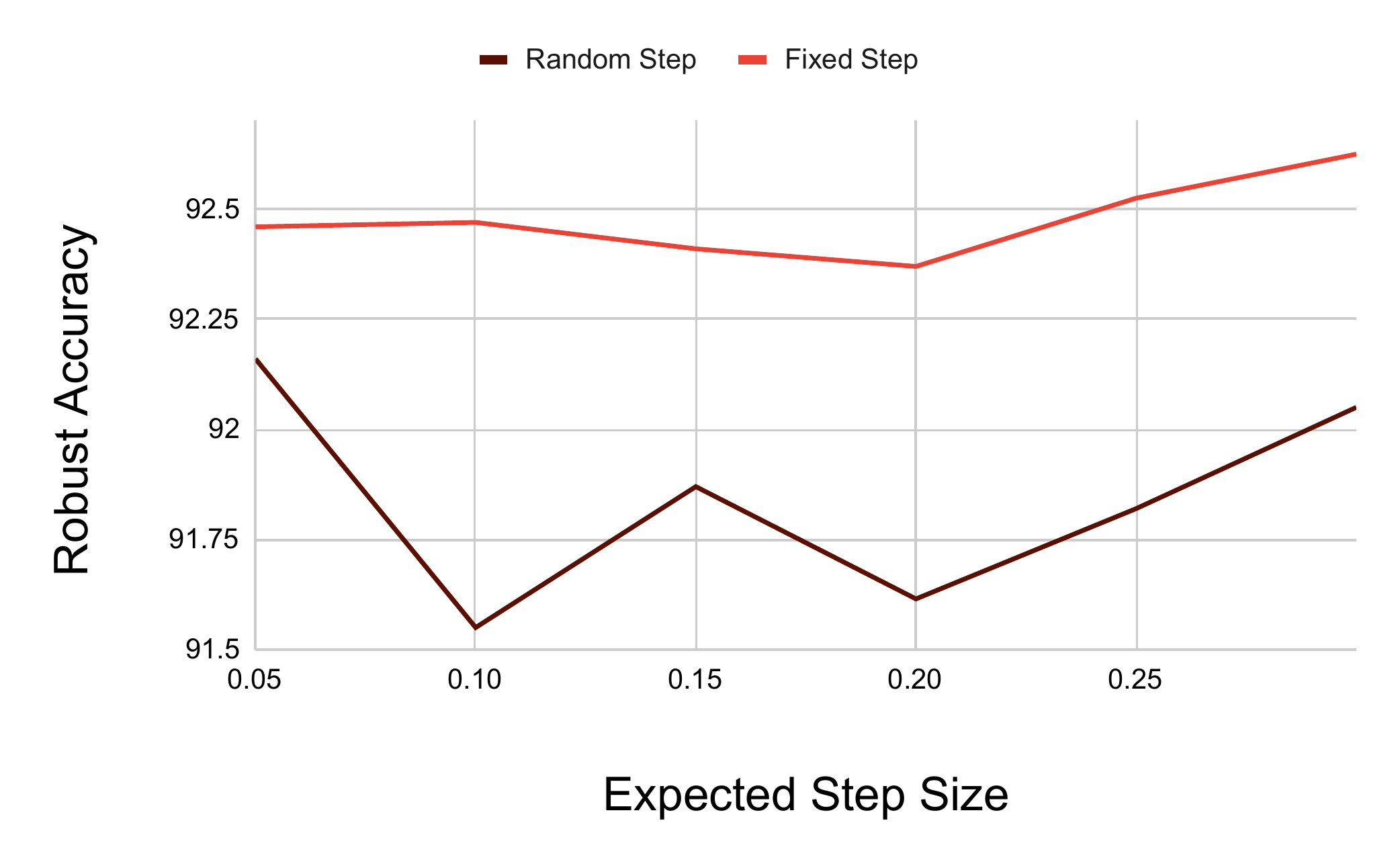}}
\caption{Sensitivity plot of a 40-PGD attack compared with 40-step WITCHcraft, this time for the more well-studied MNIST challenge.  The horizontal axis represents expected step size and the vertical axis represents robust accuracy.  As we observed above for CIFAR-10, we see that randomized step sizes result in more effective attacks against robust MNIST classifiers.}
\label{figure:sensitivity2}
\vspace{.5cm}
\end{figure}

The advantage of WITCHcraft over a range of expected step sizes is especially pronounced when attacking the difficult robust MNIST data, as Figure \ref{figure:sensitivity2} shows. We note that in this table, every value achieved by WITCHcraft surpasses any achieved by PGD.

\subsection{Exploring the benefits of additional attack steps}

\begin{figure}[h!]
{\includegraphics[width=8.5cm]{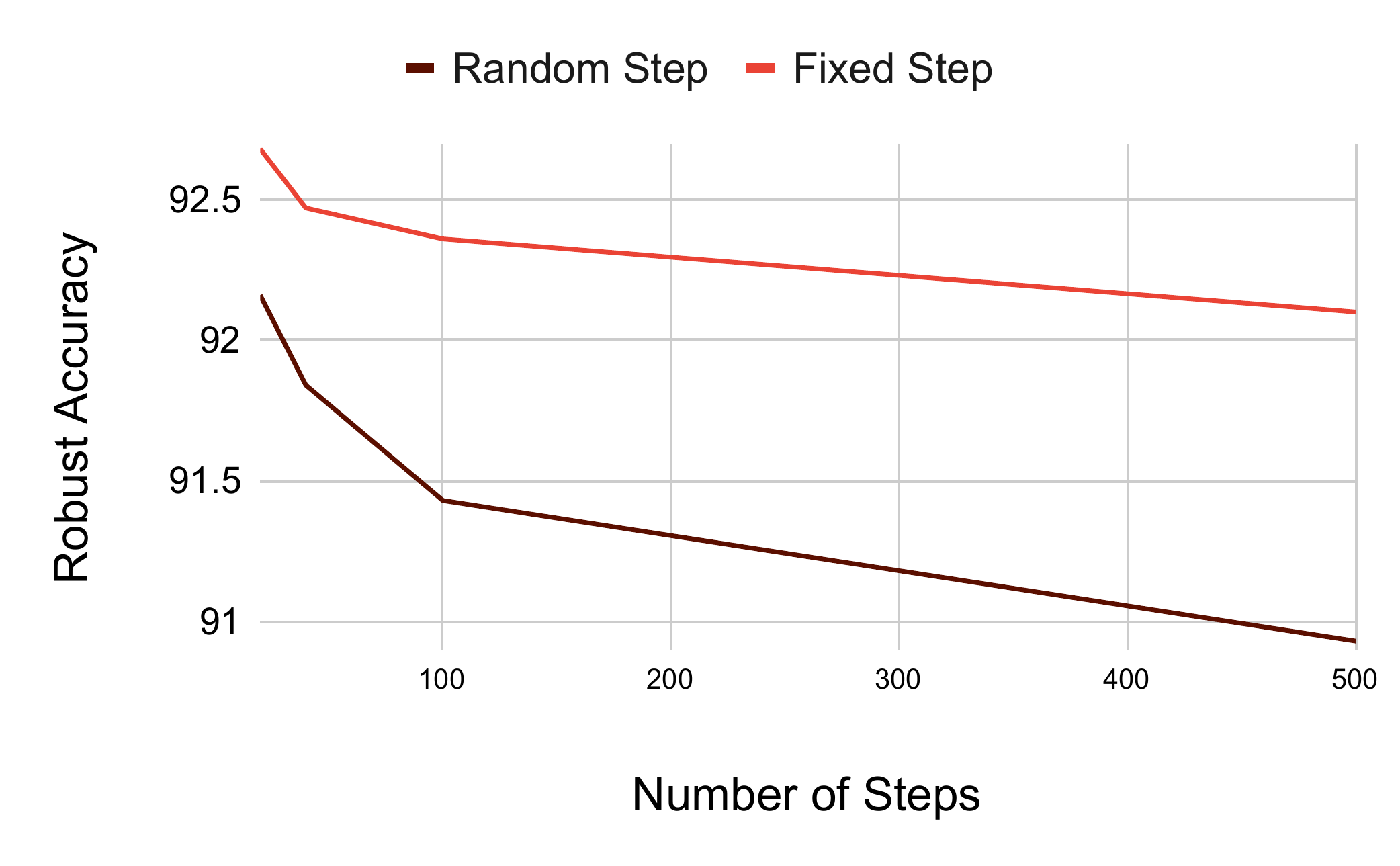}}
\caption{Comparison of robust accuracy as we increase the number of attack steps for WITCHcraft vs. PGD on MNIST. Each reported robust accuracy is an average of 6 trials. As the number of steps increases, WITCHcraft outperforms PGD.}
\label{figure:mniststeps}
\end{figure}

A third way to compare our method to PGD is to see how quickly their success rates saturate as the number of attack steps increase. Figure \ref{figure:cifarsteps} and Figure \ref{figure:mniststeps} show the results of these comparisons on CIFAR-10 and MNIST data, respectively.

We note that in both cases, WITCHcraft suffers less from diminishing returns as the number of steps grows. We hypothesize that this can be explained by the effect of randomness on exploration. The stochastic step size choice in the WITCHcraft algorithm seems to better escape local minima. The result is an algorithm that more aggressively explores the space of permissible attack images than a standard PGD attack with fixed step size.

\section{Discussion \& Conclusions}
In this work, we develop a method for introducing randomness into adversarial attacks without running the attack multiple times at different initializations. This simple modification of the popular PGD adversarial attack improves performance on benchmark data sets against robust models, while avoiding the high cost of conventional random restart methods. 
We believe that attack algorithms that perform many sequential iterations in a deterministic fashion lose efficiency due to stagnating exploration, and the WITCHcraft algorithm seems to supply a remedy for this problem.  

We hope that the proposed method can increase the efficiency of attack generation in situations like adversarial training, where attacks are crafted on-the-fly during training. A reduction in the cost of crafting attacks has the potential to make adversarial training more affordable on large industrial problems. Future work may uncover new ways to introduce randomness into attacks for increased efficiency.

\vfill
\pagebreak

\newpage
\bibliographystyle{IEEEbib}
\bibliography{refs}

\end{document}